\documentclass[sigconf]{aamas}

\usepackage{balance} 
\usepackage{array}
\usepackage{subcaption}

\setcopyright{ifaamas}
\acmConference[AAMAS '25]{Proc.\@ of the 24th International Conference
on Autonomous Agents and Multiagent Systems (AAMAS 2025)}{May 19 -- 23, 2025}
{Detroit, Michigan, USA}{A.~El~Fallah~Seghrouchni, Y.~Vorobeychik, S.~Das, A.~Nowe (eds.)}
\copyrightyear{2025}
\acmYear{2025}
\acmDOI{}
\acmPrice{}
\acmISBN{}

\title[AAMAS-2025 Formatting Instructions]{{\sf xSRL:} Safety-Aware Explainable Reinforcement Learning - \\Safety as a Product of Explainability}

\author{Risal Shahriar Shefin}
\affiliation{
  \institution{Wake Forest University}
  \city{Winston-Salem}
  \country{NC, US}}
\email{shefrs24@wfu.edu }

\author{Md Asifur Rahman}
\affiliation{
  \institution{Wake Forest University}
  \city{Winston-Salem}
  \country{NC, US}}
\email{rahmm21@wfu.edu }

\author{Thai Le}
\affiliation{
  \institution{Indiana University}
  \city{Bloomington}
  \country{IN, US}}
\email{tle@iu.edu }

\author{Sarra Alqahtani}
\affiliation{
  \institution{Wake Forest University}
  \city{Winston-Salem}
  \country{NC, US}}
\email{sarra-alqahtani@wfu.edu}

\begin{abstract}
Reinforcement learning (RL) has shown great promise in simulated environments, such as games, where failures have minimal consequences. However, the deployment of RL agents in real-world systems—such as autonomous vehicles, robotics, UAVs, and medical devices—demands a higher level of safety and transparency, particularly when facing adversarial threats. Safe RL algorithms aim to address these concerns by optimizing both task performance and safety constraints. However, errors are inevitable, and when they occur, it is essential that RL agents can explain their actions to human operators. This makes trust in the safety mechanisms of RL systems crucial for effective deployment. Explainability plays a key role in building this trust by providing clear, actionable insights into the agent’s decision-making process, ensuring that safety-critical decisions are well understood. While machine learning (ML) has seen significant advances in interpretability and visualization, explainability methods for RL remain limited. Current tools fail to address the dynamic, sequential nature of RL and its need to balance task performance with safety constraints over time. The re-purposing of traditional ML methods, such as saliency maps, is inadequate for safety-critical RL applications where mistakes can result in severe consequences. To bridge this gap, we propose xSRL, a framework that integrates both local and global explanations to provide a comprehensive understanding of RL agents' behavior. In addition, xSRL enables developers to identify policy vulnerabilities through adversarial attacks, offering tools to debug and patch agents without retraining. Thus, xSRL enhances the RL safety as a byproduct of explainability and transparency. Our experiments and user studies demonstrate xSRL's effectiveness in increasing safety in RL systems, making them more reliable and trustworthy for real-world deployment. Code is available here: \url{https://github.com/risal-shefin/xSRL}

\end{abstract}


\newcommand{\BibTeX}{\rm B\kern-.05em{\sc i\kern-.025em b}\kern-.08em\TeX}

\begin{document}


\pagestyle{fancy}
\fancyhead{}


\maketitle


\section{Introduction}\label{sec:intro}

Reinforcement learning (RL) has demonstrated significant potential in simulated environments, such as games and simplified settings, where failures have minimal consequences. However, deploying RL agents in real-world systems—such as autonomous vehicles, robotics, UAVs, and medical devices—introduces a much higher risk, as failures can lead to severe repercussions. Thus, training RL agents in these settings requires a greater emphasis on safety, especially under adversarial conditions. Addressing these concerns, the field of safe RL has emerged, developing algorithms that explicitly optimize both task performance and safety constraints \cite{kim2020safe, geibel2006reinforcement, recRL,manual_constraint, bastani2021safe, alshiekh2018safe, mihatsch2002risk,sqrl}. While these algorithms aim to enhance both safety and accuracy, errors are inevitable, and when they occur, it becomes crucial for the agent to explain its behavior to human practitioners. Practitioners must discern whether an agent is behaving correctly, or is malfunctioning and thus requires intervention. Therefore, user trust in these safety approaches is vital; without it, practitioners may disregard the systems, undermining their effectiveness. Explainability fosters this trust by allowing practitioners to query the reasoning behind safety-related decisions, alongside the agent's learned policy. This additional transparency helps make safety decisions more actionable and strengthens practitioners' confidence in the RL system. In this work, we hypothesize that \textit{safety in RL is intrinsically tied to explainability and transparency}, which allow the users to gain a clear understanding of agent behavior, enable efficient debugging and address its safety concerns.

\textbf{The need for explainable RL (XRL).} In machine learning (ML), significant progress has been made in developing interpretation and visualization methods to help users understand a ML model's performance, track its metrics \cite{track}, generate data-flow graphs for its decision-making \cite{graphs}, and visualize representations it has learned \cite{embedding}. However, interpretation tools designed specifically for RL agents are still limited, most of which only offer basic capabilities such as behavior summarization \cite{summarize}, contrastive explanations \cite{contrastive}, or short video clips at critical time steps, but without providing any in-depth reasoning about the RL agents' behavior. While re-purposing ML interpretation methods (e.g., saliency maps~\cite{Simonyan2013DeepIC}) might seem feasible, their design intent is insufficient for explaining RL agents, especially in safety-critical environments. RL agents must not only optimize task performance but also adhere to strict safety constraints, as any mistake can lead to significant harm to agents or human workers. Existing ML explainability tools focus on individual, static decisions, which fails to capture the sequential and dynamic nature of RL, where actions impact future states and they require balancing immediate rewards with long-term risks. Additionally, RL agents often operate under stochastic policies and delayed feedback, making it difficult to explain how they learn to manage the trade-off between exploration and exploitation while maintaining safety. Therefore, safety-aware explainability must address both task performance and the agent’s adherence to safety constraints, ensuring that every action is evaluated in terms of both its short- and long-term benefits and risks—a complexity that re-purposed ML tools are not well-equipped to handle.

\textbf{Desiderata.} Effective XRL methods for safety must meet key criteria to ensure usability and trustworthiness. First, they must provide global explanations of the agent's policy, offering insights into its behavior across the state space \cite{McCalmon,Hayes2017ImprovingRC,zahavy2017graying,TopinV19}, especially for end-users unfamiliar with RL. Second, they should deliver local explanations, detailing the agent’s reasoning at specific timesteps. Third, explanations must be interpretable, preferably in natural language, with minimal user effort. Fourth, the method must ensure explanation fidelity, validated through tests \cite{Markus_2021} to confirm alignment with the agent’s actual policy. Finally, XRL should include adversarial explainability, analyzing how decisions change under adversarial conditions to identify vulnerabilities and improve robustness. Beyond user trust, XRL should provide utilities for RL users and developers, such as debugging, testing, and patching policies to enhance safety and performance \cite{gain,Hohman,Mohseni2018ASO}. Currently, no tools or frameworks address all these needs.

\textbf{Proposal.} To address the desiderata, we propose xSRL, a novel framework integrating local and global explanations to enhance RL safety through transparency and explainability. xSRL explains safety violations by showing how task requirements or safety constraints influence decisions. xSRL introduces a novel local explanation method, training two critics, $Q_{task}$ and $Q_{risk}$, to estimate returns and risks for a state $s_t$, action $a_t$, and policy $\pi$. These post-hoc Q-functions apply to any fixed policy without accessing its internal structure. For global explanations, xSRL extends our previous work, CAPS \cite{McCalmon}, enhancing nodes (abstract states) with $Q_{task}$ and $Q_{risk}$, and annotating edges (actions) with motivations rooted in task or safety requirements.
xSRL also provides adversarial explainability, enabling developers to debug and patch policies by identifying vulnerabilities and analyzing behavior under risk. Developers can launch adversarial attacks and improve safety without retraining the policy. To our knowledge, xSRL is the first framework combining local and global explanations to address RL agent safety and the first to offer adversarial explanations for vulnerability analysis and policy patching.

\textbf{Key Contributions.} We show that xSRL' explanations provide the users with both \textit{trust} and \textit{utility} to understand and protect RL agents in critical scenarios.
Trust is measured computationally through fidelity tests and empirically via user studies. Fidelity tests assess the alignment between the explanation graphs and the agent's actual policy across two benchmark environments in safe RL \cite{recRL, adv}, using safety techniques \cite{adv,recRL,sqrl} to patch vulnerable policies. Results indicate that xSRL generates accurate policy graphs with less than 33.5\% error in risk estimation, even under adversarial attacks. To further validate xSRL’s effectiveness, we compared xSRL against local and global explanations by evaluating users’ comprehension of xSRL explanations for both vulnerable and safe policies under adversarial attacks in a user study with 9 distinct conditions. These conditions represented various combinations of local and global explanations with different safe RL patching methods. Both xSRL and global explanations (CAPS \cite{McCalmon}) achieved comparable accuracy, demonstrating the superiority of global explanations (xSRL and CAPS) over local explanations in providing clearer insights in high-risk contexts. However, xSRL offers a distinct advantage over CAPS by facilitating debugging and patching for safety constraints. We also evaluated the impact of explanation-guided adversarial attacks on an agent trained with the soft-actor-critic (SAC) algorithm \cite{sac}, where the attack reduced the agent’s safety by approximately 72\% under a 50\% attack rate. After identifying vulnerabilities in the SAC agent's policy, we patched it with a safety shield \cite{shield1} and a safe policy, significantly improving the agent's safety. Lastly, user studies assessed participants' ability to distinguish the safer agent between two alternatives, confirming xSRL’s utility in enhancing safety and providing actionable insights. Overall, xSRL effectively increases the trustworthines and utility of RL agents for real-world, safety-critical applications by identifying and resolving policy vulnerabilities.

\section{Related Work}
XRL methods have evolved to offer insights into the behavior of RL agents, with approaches generally categorized into local and global explanations \cite{molnar2022}. \textit{Local explanations} focus on specific actions taken by an agent at a point in time, often using post-hoc methods like saliency maps \cite{Hilton2020UnderstandingRV,Huber2021BenchmarkingPS,puri}, which highlight key features that influence the decisions. Intrinsic methods, such as reward decomposition \cite{Juozapaitis2019ExplainableRL}, build explainability into the agent's decision model itself by breaking down the Q-value into components that reveal the agent's motivations at each time step. These approaches help in understanding why specific decisions are made but do not provide insight into the satisfaction or violation of safety constraints explicitly.

Global explanations attempt to describe an agent's overall strategy by summarizing its behavior across various states. Recent work introduced the concept of ``agent strategy summarization'' \cite{Amir2019,summarize}, where an agent’s behavior is demonstrated through its actions in a carefully chosen set of states. The key challenge in this approach is how to select the most crucial state-action pairs that effectively portray the agent’s behavior, allowing users to anticipate how it may act in new scenarios. Techniques such as HIGHLIGHTS \cite{summarize} identify important states based on the impact of decisions on the agent's utility, while other methods use machine learning to optimize state selection \cite{enable,Lage}. These summaries reduce the effort required for humans to understand the agent’s behavior while still providing comprehensive information about its capabilities.

Combining local and global methods, such as integrating strategy summaries with saliency maps \cite{Huber2021BenchmarkingPS}, has shown promise but lacks the depth needed for understanding safety-oriented decisions such as the estimation of the risk of agent's failing at each state and for the overall policy. In general, none of the existing XRL methods explicitly explain how an agent adheres to or violates safety constraints.  Moreover, current XRL approaches primarily focus on increasing end-user trust by providing interpretable models that emphasize behavior, without evaluating how these explanations could benefit developers tasked with debugging or refining the RL agents.

Our work addresses these gaps by developing xSRL, a framework that integrates local and global explanations specifically designed to convey the satisfaction or violation of safety constraints explicitly. xSRL combines strategy summarization with safety-focused local explanations, illustrating how agents balance task performance with risk assessments across different states. We hypothesize that \textit{"safety is a product of explainability"} by offering actionable insights that enable developers to interactively debug and refine RL policies, thereby enhancing safety. This utility is crucial for improving RL policies in safety-critical environments, where errors can have severe consequences. \textit{To our knowledge, no existing XRL method provides such a comprehensive, safety-oriented approach, highlighting the importance and novelty of our approach.}

\section{Background} \label{backgroung}

For the purpose of explaining safety of RL agents, we consider the standard Constrained Markov Decision Processes (CMDPs), \(\mathcal{M}=(\mathcal{S},\mathcal{A},\mu,\mathcal{P(.|.,.)},\mathcal{R},\gamma, \mathcal{C}) \) where \(\mathcal{S}\) and \(\mathcal{A}\) denote the state and action space; \(\mu\) and  \(\mathcal{P}:\mathcal{S}\times  \mathcal{A}\times\mathcal{S}\xrightarrow{}[0,1] \) denote the initial state distribution and state transition dynamics, respectively.  \(\mathcal{R}: \mathcal{S}\times \mathcal{A}\xrightarrow{}\mathcal{R}\) is the reward function; \(\gamma \) denotes discount factor; and \(\mathcal{C}=\{ c_{i}: \mathcal{S}\times \mathcal{A}\xrightarrow{\mathcal{S}}\mathbb{R}\geq0;\:i=1,2,..,T\} \) denotes the set of cost associated with safety constraint violations in any trajectory episode \(\tau=\{s_0, a_0,..., a_{T-1},s_T\}\) with a maximum trajectory length of \(T\). We assume that either accomplishing the task goal or violating a safety constraint in \(\mathcal{M}\) leads to episode termination. The objective of the agent's policy, we call it \textit{task policy} hereafter, \(\pi_{task}\) is to learn the optimal control to maximize the expected discounted reward at time \(t\); \(\mathcal{R}_{\pi_{task}}=\mathbb{E}_{\tau \sim \pi_{task}}[\sum_{t^{'}=t}^{T}\gamma^{t^{'}-t}r_t]\). The task \emph{policy} $\pi_{task}$ is the solution to the CMDP.

\textbf{Problem Statement.} Our ultimate goal is to provide a comprehensive explanation of an RL agent's behavior. Consider an RL task solved by an agent trained with either value-based algorithms such as DQN \cite{DQN} and DDQN \cite{ddqn} or policy-based algorithms such as PPO \cite{ppo} or TRPO \cite{trpo}. This paper aims to explain this agent's policy by summarizing its overall strategy to solve the task. Formally, given $N$ episodes $\mathbb{T}=\{\mathbf{X}^{(i)}, r_{i}, c_{i} \}_{i=1:N}$ of the target agent, $\mathbf{X}^{(i)}=\{s^{(i)}_{t},a^{(i)}_{t},r^{(i)}_{t},c^{(i)}_{t}\}_{t=1:T}$ is the i-th episode of length $T$, where $s^{(i)}_{t}$ is the state, $a^{(i)}_{t}$ is the action, $r^{(i)}_{t}$ is the reward, and $c^{(i)}_{t}$ is the cost, at time $t$ in episode $i$. Our goal is to generate a summary of these episodes as a \textit{graph} accompanied by estimated values of Q-functions for the \textit{task reward} and the \textit{cost of safety violations}.

\textbf{Baseline.} In this paper, we build on our previous method, CAPS \cite{McCalmon}, a recently introduced global explanation XRL method that has been successful in providing comprehensible Summaries of RL policies, as our main comparison baseline. CAPS collects natural language (NL) predicates from the user and gathers up to 500 timesteps from the RL agent’s trajectories. To simplify the explanation process, it uses the CLTree \cite{cltree} clustering algorithm to abstract the agent’s states into clusters, forming a hierarchy. A heuristic optimization then selects the best cluster configuration based on state transition accuracy and user interpretability. CAPS constructs the agent’s policy ($\pi$) as a directed graph $G=(V,E)$, where set of nodes $V$ represent abstract states (clusters), and set of edges $E$ show the agent’s actions and transition probabilities. CAPS enriches the graph by labeling abstract states with English explanations based on user-defined predicates and boolean algebra. We chose CAPS \cite{McCalmon} because it offers a comprehensive global explanation, revealing the agent’s policy across the state space rather than focusing on specific states. It also outperforms other global XRL methods \cite{TopinV19, zahavy2017graying} in fidelity and user comprehension.

While CAPS \cite{McCalmon} can explain an RL agent’s policy, it cannot address the impact of safety violations on the agent's behavior, nor does it enable users to debug specific safety concerns. \textit{xSRL addresses these limitations by providing safety through enhanced explainability and transparency.}

\section {Approach: Safety-Aware Explainable RL Method}
 In this section, we present our approach, so-called xSRL, which uniquely combines a novel local explanation method with an extended version of the global explanation framework CAPS~\cite{McCalmon} to explain an RL agent’s decision-making process at both individual states and its overall strategy to achieve a comprehensive consideration for both task and safety requirements.

\subsection{Safety Interpretation via Integrating Local and Global Explanations} \label{integration}
\noindent \textbf{Local Explanation Method.} In XRL, reward decomposition \cite{Juozapaitis2019ExplainableRL} is used to reveal the reasoning behind an agent’s actions in specific states by decomposing the reward into different components. This method, which separates rewards into individual reward components values $R_{c}(s,a)$, highlights the factors influencing an agent's decisions at each timestep. However, such approach is not built to explain safety for RL agents, which can be optimized in a separate objective function through joint optimization \cite{sqrl,thananjeyan2020safety,mihatsch2002risk} or provided through a separate policy from the task policy \cite{adv,recRL}. To bridge this gap, we propose a local explanation method that can explicitly explain safety in RL. Our local explanation method consists of 2 components: task reward estimation and risk estimation at each ($s,a$) pair.

 To estimate the future risk probability of a safety-constrained task policy $\pi_{\mathrm{task}}$, we train a separate risk-critic $Q_{\mathrm{risk}}$ that evaluates the safety of a given state-action pair independently of task objectives. This risk-critic function $Q^{\pi_{\mathrm{task}}}_{\mathrm{risk}}(s,a)$, learned via:

\begin{equation}\label{q_risk} Q^{\pi_{\mathrm{task}}}_{\mathrm{risk}}(s,a)=\mathbb{E}_{{a_{t}\sim \pi_{\mathrm{task}}(.|s_{t})}}[\sum_{t^{'}=t}^{T}\gamma_{risk}^{t^{'}-t}c(s_t,a_t)], \end{equation}

\noindent provides a way to assess the safety constraints of actions taken by the task policy. Here, $c(s_t, a_t)$ denotes the cost associated with violating safety constraints when taking action $a_t$ in state $s_t$. In practice, we approximate $\hat{Q}^{\pi_{\mathrm{task}}}_{\phi,\mathrm{risk}}$, parameterized by $\phi$, using sampled transitions $(s_{t},a_{t},s_{t+1}, c_{t})$. This is done by minimizing the following MSE loss with respect to the target (RHS of Eq. \ref{q_risk}):
\begin{equation}
\begin{split}
    J_{\mathrm{risk}}(s_{t},a_{t},s_{t+1};\phi) &= \frac{1}{2} \left(\hat{Q}^{\pi_{\mathrm{task}}}_{\phi,\mathrm{risk}}(s_{t},a_{t})  - \right. \\
    & \left .\mathbb{E}_{a_{t+1} \sim \pi(.|s_{t+1})} [\hat{Q}^{\pi_{\mathrm{task}}}_{\phi,\mathrm{risk}}(s_{t+1},a_{t+1})] \right)^{2}
\end{split}
\end{equation}

To estimate the future task reward for a task policy $\pi_{\mathrm{task}}$ at each state, we train a separate task-critic $Q_{\mathrm{task}}$ similar to $Q_{\mathrm{risk}}$ in Eq.\ref{q_risk} but for the task reward $r$ instead of safety cost $c$:
\begin{equation}\label{q_task} Q^{\pi_{\mathrm{task}}}_{\mathrm{task}}(s,a)=\mathbb{E}_{a_{t} \sim \pi_{\mathrm{task}}(.|s_{t})}[\sum_{t^{'}=t}^{T}\gamma_{\mathrm{task}}^{t^{'}-t}r(s_t,a_t)] \end{equation}

Since value-based RL algorithms, in general, and actor-critic policy-based algorithms estimate Q-function for the task policy, we can directly utilize their $Q_{\mathrm{task}}$ to explain their performance at each state in terms of task requirements.

\vspace{3pt}
\noindent \textbf{Global Explanation Method.} To enhance the effectiveness of local explanations, we propose integrating both $Q_{\mathrm{task}}$ and $Q_{\mathrm{risk}}$ into the global explanation method, CAPS \cite{McCalmon}. This approach allows users to understand how the agent balances task objectives with safety constraints while presenting its overall strategy across episodes. CAPS summarizes the agent's policy in a directed graph where nodes represent abstract states and edges represent actions along with their transition probabilities, as described in Section \ref{backgroung}. We choose to present xSRL explanations in a directed graph format because it effectively illustrates the relationships and dependencies among states, actions, and outcomes; their causal and safety relationships; and the progression from one state to another based on specific actions. However, using CAPS alone as a global explanation tool did not provide sufficient insight for users to determine why the agent chose a particular action at a specific state, as demonstrated in our user studies (Section \ref{evaluation}).

Our method xSRL improves CAPS graph by: (1) incorporating local $Q_{\mathrm{task}}$ and $Q_{\mathrm{risk}}$ to show task and risk estimation at each abstract state, and (2) explicitly indicating whether each action is driven by task or safety considerations. We hypothesize that combining global policy summaries with Q-function decomposition will significantly enhance user understanding of agent safety, compared to relying solely on local or global explanations, as shown in our evaluation section. To compute $Q_{\mathrm{risk}}(B,a)$ and $Q_{\mathrm{task}}(B,a)$ for an abstract state $B$ and action $a$, we average $Q_{\mathrm{risk}}(s,a)$ and $Q_{\mathrm{task}}(s,a)$ over all concrete states $s \in B$ and all possible actions $a$ from $s$:
\begin{equation}\label{Q-task}
\begin{split} Q_{\mathrm{task}}^{\pi_{\mathrm{task}}}(B,a)= \frac{1}{n} \mathbb{E}_{a \sim \pi(.|s)} \sum_{i=1}^{n} Q^{\pi_{\mathrm{task}}}_{\mathrm{task}}(s,a)
\end{split}
\end{equation}

\begin{equation} \label{Q-risk}
\begin{split} Q_{\mathrm{risk}}^{\pi_{\mathrm{task}}}(B,a)= \frac{1}{n} \mathbb{E}_{a \sim \pi(.|s)} \sum_{i=1}^{n} Q^{\pi_{\mathrm{task}}}_{\mathrm{risk}}(s,a),
\end{split}
\end{equation}

\noindent where $n$ represents the total number of concrete states within the abstract state $B$. These values are attached to the abstract state $B$ in the directed graph, showing how the agent's policy $\pi$ evaluates task satisfaction (Eq.\ref{Q-task}) and safety constraint violations (Eq.\ref{Q-risk}) when taking action $a$ from state $B$. The second enhancement we added to CAPS will be discussed in Section \ref{patching}.

\subsection{Safety Debugging via Adversarial Explanation} \label{guided-attack}


This section illustrates how xSRL can be used to provide so-called \textit{adversarial explanations} to help users in discovering and debugging vulnerabilities in RL policies. Specifically, we demonstrate how users, using the information revealed by xSRL, can launch adversarial attacks to explore potential pitfalls of an RL agent. Through xSRL's graphs, users can also explain the agent’s mistakes and its violations of safety constraints, allowing them to formulate a remediation policy that improves the agent's original behavior (discussed in Section \ref{patching}).

To initiate an adversarial attack, we first collect 500 episodes ($\tau$) from the target agent and explain them using xSRL's graph $\mathbb{G}$. Next, using $\mathbb{G}$, we identify the top-K safety-critical states across all episodes, defined as follows:

\noindent \textbf{Definition 1: Safety-Critical State}. \textit{Given a policy $\pi_{\mathrm{task}}$, a state $s_c$ is a safety-critical state iff there is at least one action $a$ chosen by $\pi_{\mathrm{task}}$ such that:}
\begin{equation} \label{critical}
    Q^{\pi_{\mathrm{task}}}_\mathrm{{risk}}(s_{c},a) > \epsilon_{\mathrm{safety}},
\end{equation}
\noindent where \textit{the set of all safety-critical states $\mathbb C$ is $\forall s_{c} \in S$.} Finally, we run the agent for another 500 episodes, forcing it to take adversarial actions at the common critical states $ \mathbb C$ identified by $\mathbb{G}$ at varying rates (as we show in Section \ref{evaluation}), while collecting its trajectories $\tau_{A}$. These adversarial actions are generated using the Alternative Adversarial Action (AAA) attack \cite{act_perturb}, employing a pre-trained adversarial policy \(\pi^{\mathrm{adv}}\) from \cite{adv} to select alternative adversarial actions \(a_{\mathrm{adv}}\sim \pi^{\mathrm{adv}}(s_t)\). We then use xSRL to generate explanation graph for the agent under attack $\mathbb{G_{A}}$, based on $\tau_{A}$. By contrasting both graphs, $\mathbb{G}$ and $\mathbb{G_{A}}$ without and with the adversarial attacks, respectively, users can pinpoint the safety vulnerabilities in the agent's policy by noticing the overall graphs and their attached $Q_{\mathrm{risk}}$ and $Q_{\mathrm{task}}$ values.

\subsection{Patching Explanation-Based Discovered Vulnerabilities} \label{patching}

This section demonstrates how xSRL can guide the patching process for vulnerabilities discovered in RL policies through adversarial explanations. To avoid retraining the task policy $\pi_{\mathrm{task}}$, we propose a simple approach using an auxiliary policy that optimizes only the safety violation cost function $c$. Specifically, we employ the safety policy $\pi_{\mathrm{safety}}$ from \cite{adv}, which is trained by maximizing the KL-divergence from an adversarial policy that increases safety violation costs. We then adopt the post-posed shielding strategy from \cite{shield1} during online execution to shield the states with higher $Q_{\mathrm{risk}}$ from attacks. To implement the safety shield, we leverage the risk-critic $Q_{\mathrm{risk}}$  trained for explainability in (Eq.\ref{q_risk}) as:
\begin{equation}
    \begin{aligned} \label{shield}
     Shield(s_t,a_t) &: Q_{\mathrm{risk}}(s_t,a_t)>\mathbb{T}_{\mathrm{safety}},
    \end{aligned}
\end{equation}
where \(\mathbb{T}_{\mathrm{safety}}\) is a predefined threshold value such that at any state \(s_t\) and for any action \(a_t \sim \pi^{\mathrm{task}}(s_t)\); if \(Shield(s_t,a_t)\) is triggered, then the shield replaces the selected action \(a_t\) by a safer action given by the safety policy \(a^{\mathrm{safe}}_{t}\sim \pi^{\mathrm{safety}}(s_t)\). The value of \(\mathbb{T}_{\mathrm{safety}}\) is environment-specific and can be chosen based on a sensitivity test for each environment.

xSRL can then be used to generate an updated graph for the patched agent, enabling developers to assess improvements or uncover new vulnerabilities. Since the proposed patching process, along with certain safe RL algorithms like \cite{adv,recRL}, involves two distinct policies—task and safety, it provides additional data to further refine the xSRL graphs. Using the shielding threshold, each edge $(B,a)$ in the graph is labeled as a ``safety decision'' if action $a$ is chosen by $\pi_{\mathrm{safety}}$, or a ``task decision'' if selected by $\pi_{\mathrm{task}}$. The responsible policy is identified by tracking which policy (safety or task) selects the actions in each concrete state $s \in B$, with the dominant policy making the most decisions in $B$ being assigned.

\begin{figure*}[t] 
    \centering
    \begin{subfigure}{0.32\textwidth} 
        \includegraphics[width=\linewidth]{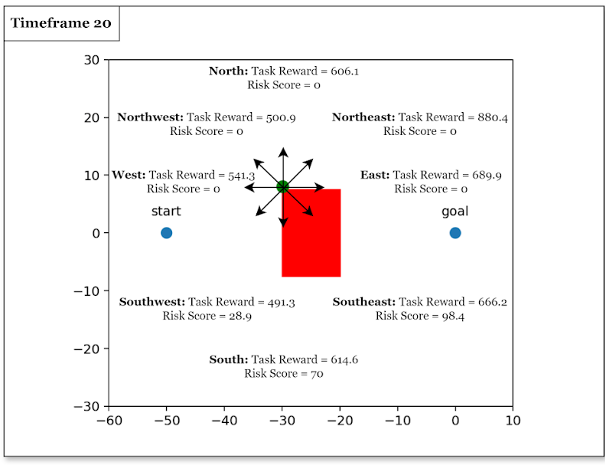} 
        \caption{Local explanation}
        \label{fig:subfig1}
    \end{subfigure}
    \hfill
    \begin{subfigure}{0.32\textwidth}
        \includegraphics[width=\linewidth]{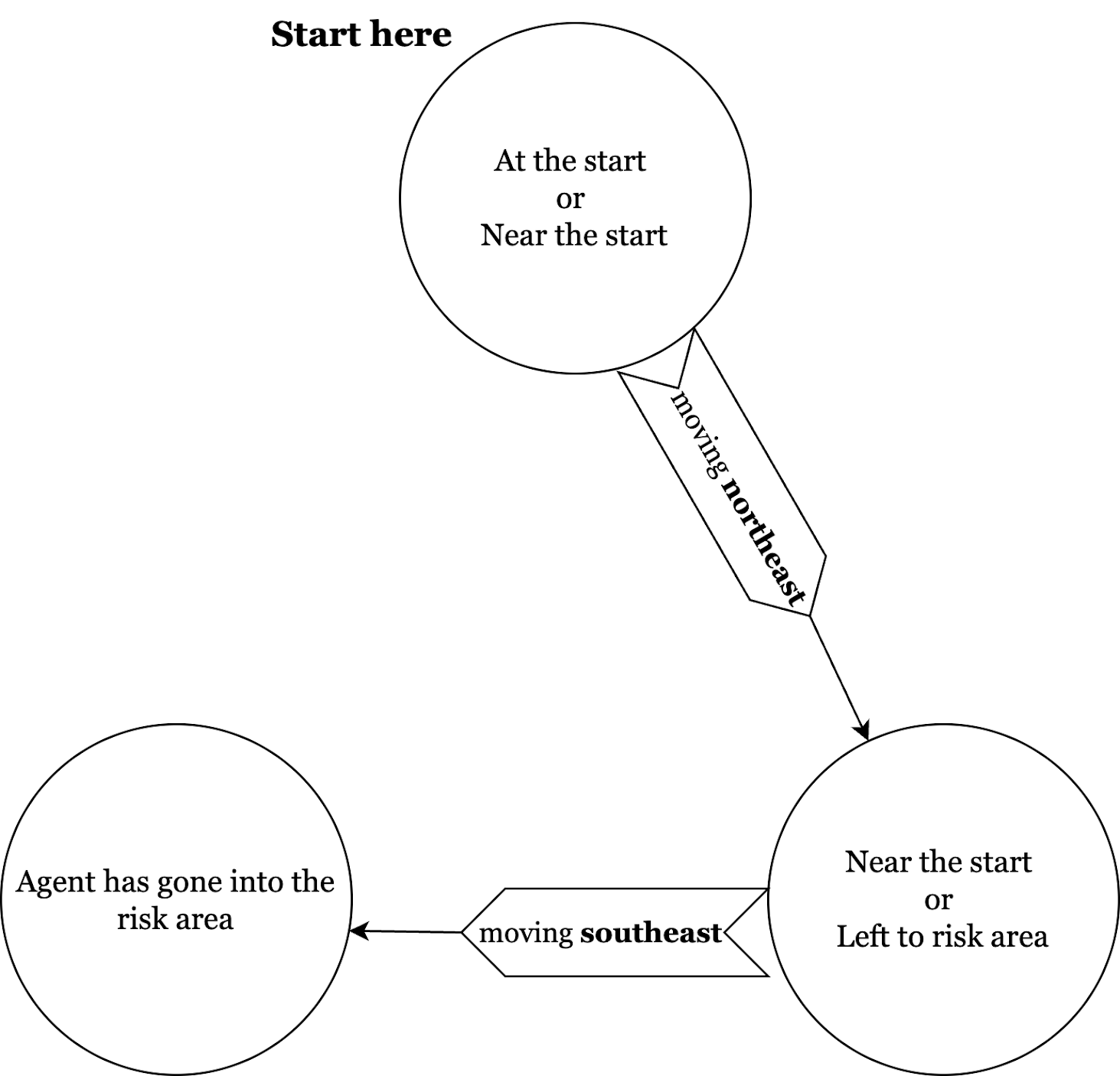} 
        \caption{Global explanation}
        \label{fig:subfig2}
    \end{subfigure}
    \hfill
    \begin{subfigure}{0.32\textwidth}
        \includegraphics[width=\linewidth]{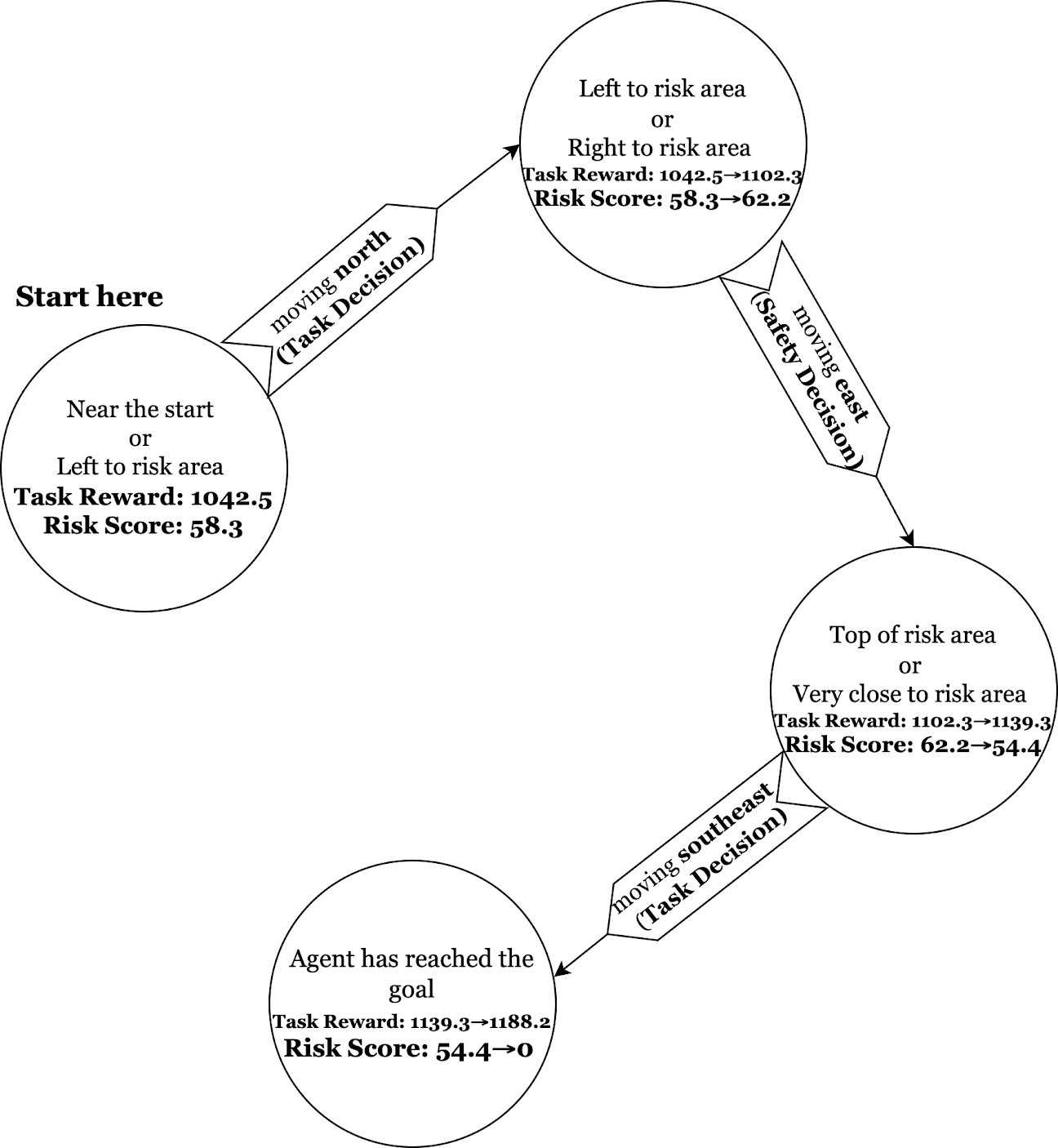} 
        \caption{xSRL (local and global) explanation}
        \label{fig:subfig3}
    \end{subfigure}

    \caption{Examples of generated explanations for Navigation2 task using our local explanation, global explanation from CAPS \cite{McCalmon}, and our xSRL that integrates local and global explanations.}
    \label{fig:examples}
\end{figure*}

\section{Evaluation} \label{evaluation}
In this section, we evaluate xSRL based on two key objectives: trust and utility. Trust assesses whether the explanations generated by xSRL are comprehensible to end-users, while utility measures the effectiveness of these explanations in identifying and resolving vulnerabilities in the agent’s policy.

\vspace{3px} \noindent \textbf{Tasks:} We conducted our experiments and user studies in two continuous MuJoCo CMDP environments \cite{recRL}: (1) Navigation 2, and (2) Maze. In both environments, the state-actions spaces are continuous and the agent's objective is to reach a goal state while avoiding collisions with obstacles, walls, or boundaries. For each task, we used a well-trained SAC agent \cite{sac} as the target. We present our results for Navigation 2 in the main paper, with \textit{the results for Maze provided in Appendix D}.

 \vspace{3px} \noindent \textbf{Explanation Baselines:} As discussed in Section \ref{integration}, XRL methods can be classified into two broad categories: (1) local explanations and (2) global explanations.
 For comparison, we select two representative baselines. The first baseline is our local explanation method, which presents users with $Q_{\mathrm{task}}$ and $Q_{\mathrm{safety}}$ values at specific time steps. The second baseline is CAPS\cite{McCalmon}, which generates a directed graph summarizing the agent’s overall policy. Figure.\ref{fig:examples} illustrates three different explanations in Navigation2 using our local explanation method, global explanation (CAPS), and xSRL, which integrates both local and global explanations, for an agent patched with the safe policy from \cite{adv}.

\vspace{3px} \noindent \textbf{Safety Patching Baselines:} We employed three different safe RL methods as patching techniques. First, we used the safety policy $\pi_{\mathrm{safe}}$ developed in AdvExRL \cite{adv} and RRL-MF \cite{recRL}, separately. These methods represent approaches that optimize safety separately from task performance. Second, we tested SQRL \cite{sqrl}, which exemplifies joint optimization of both task performance and safety. Details of these methods are given in Appendix A.

\subsection{Trustworthiness of xSRL's Explanations}\label{sec:trust}
We evaluate the trustworthiness of xSRL explanations using two approaches: computational fidelity scores and user studies.

\subsubsection{Fidelity} \leavevmode\newline
\noindent \textbf{Overview.} Fidelity measures how accurately the produced explanation reflect the true behavior of the RL agents. The higher the fidelity, the better alignment between the graph and the agent's behavior, the better the explanation. We calculate fidelity for four components of the agent that xSRL explanations capture: (i) action, (ii) policy selection, (iii) $Q_{\mathrm{risk}}$, and (iv) $Q_{\mathrm{task}}$. The \textit{action} fidelity is measured by the proportion of actions taken by the RL agent that match the \textit{actions} predicted by the produced explanation graph \cite{Markus_2021}. Similarly, policy selection fidelity measures the extent to which the graph correctly identifies the policy used by the agent to make decisions. $Q_{\mathrm{risk}}$ and $Q_{\mathrm{task}}$ fidelities measure how closely xSRL's local explanations match the values produced by the agent's policy. We use the weighted average of the Normalized Root Mean Squared Error (NRMSE) to compute the difference between the $Q_{\mathrm{risk}}$ and $Q_{\mathrm{task}}$ estimates generated by xSRL and those from the agent’s policy. A lower NRMSE indicates a better explanation.

To compute all the fidelity scores, we simulate around 2,000 timesteps of agent-environment interactions for several agents: SAC \cite{sac} without attacks, SAC under a 50\% attack, SAC patched with the safe policy $\pi_{\mathrm{safety}}$ from AdvExRL \cite{adv}, the safe policy from RRL-MF \cite{recRL}, and the SQRL \cite{sqrl}-trained agent. We generated and averaged fidelity scores across five graphs for each agent.

\begin{table}[ht]
\centering
\caption{Fidelity scores for the explanations generated by xSRL for different agents: SAC \cite{sac} (with and without attacks), and agents patched with AdvExRL \cite{adv}, RRL-MF \cite{recRL}, and SQRL \cite{sqrl}. A higher action and policy selection ratio indicates better explanation (policy selection is only applicable for AdvExRL \cite{adv} and RRL-MF \cite{recRL}). For $Q_{\mathrm{risk}}$ and $Q_{\mathrm{task}}$, a lower NRMSE indicates better fidelity.}
\begin{tabular}{lcccc}
\toprule
 & \textbf{Action} & \textbf{Policy Selection} & \textbf{$Q_{\mathrm{risk}}$} & \textbf{$Q_{\mathrm{task}}$} \\
\midrule
\textbf{SAC (w/o attack)}  & \textbf{63.4\%} & - & 44.5\% & 30.4\% \\
\textbf{SAC (w/ attack)}    & 42.8\% & - & 43.1\% & 39.2\% \\
\textbf{advExRL}            & 34.25\% & 59.2\% & 33.2\% & 32.6\% \\
\textbf{RRL-MF}             & 31.7\% & \textbf{94.7\%} & 48.4\% & 27.4\% \\
\textbf{SQRL}               & 32.4\% & - & 62.8\% & 25.6\% \\
\bottomrule
\end{tabular}\label{fidelity-scores}
\end{table}

\noindent \textbf{Results and Discussion.} Table~\ref{fidelity-scores} summarizes the results and highlights several important trends. The SAC agent without attack achieves the highest action ratio fidelity (63.4\%), demonstrating that xSRL effectively captures the agent's behavior in a stable environment. However, under attack, the action ratio fidelity for SAC drops to 42.8\%, reflecting the agent's reduced consistency under adversarial conditions, which also impacts the accuracy of the explanations. Among the patched agents, AdvExRL performs notably well with a relatively high policy selection ratio (59.2\%) and the lowest $Q_{\mathrm{risk}}$ NRMSE (33.2\%), indicating that AdvExRL provides the most accurate safety explanations. This can be attributed to AdvExRL’s training, which maximizes the KL divergence from an optimal adversarial policy \cite{adv}, making its safety policy more optimal and, therefore, more explainable. RRL-MF achieves the highest policy selection fidelity (94.7\%) due to its lower risk estimation, where most actions were taken by the task policy, as reflected in its xSRL graphs (see Appendix C). Confirming that, RRL-MF’s $Q_{\mathrm{risk}}$ NRMSE is relatively high (48.4\%), indicating challenges in estimating risk. This is likely because RRL-MF relies on expert demonstrations for training safe behavior \cite{recRL}, which may not cover all possible risky scenarios, leading to less accurate safety explanations. On the other hand, SQRL excels in task-related explanations with the lowest $Q_{\mathrm{task}}$ NRMSE (25.6\%), but it has the highest $Q_{\mathrm{risk}}$ NRMSE (62.8\%). This discrepancy in risk estimation is likely due to SQRL’s joint optimization of both task and safety requirements \cite{sqrl}, making it harder to isolate and accurately estimate risk values.

Overall, xSRL provides comprehensive explanations, but its local safety explanations ($Q_{\mathrm{risk}}$) are highly influenced by the safety patching mechanism used. Thus, we recommend using separate policies for safety, as seen in AdvExRL \cite{adv}, which yields more optimal and interpretable safety explanations.

\subsubsection{User Studies}  \leavevmode\newline
\noindent \textbf{Overview.} To empirically assess users' trust in xSRL explanations, we conducted three user studies to evaluate the impact of combining local and global explanations in xSRL, as well as the effect of each of two levels of explanation individually. Each study used three different patched agents (AdvExRL, RRL-MF, and SQRL), resulting in a total of nine distinct conditions. As local explanations apply to specific states, we selected states with the highest $Q_{\mathrm{risk}}$ to show users the agent's behavior in risky scenarios.

\vspace{3pt}
\noindent \textbf{Procedure.} We recruited 270 participants (30 per study), consisting of sophomores from various majors at our institution and participants from Prolific, with IRB approval. We applied specific filters to ensure relevant participant backgrounds. Each participant was randomly assigned to one study to avoid crossover effects. Initially, participants were introduced to Navigation 2 environment, followed by an explanation of the safety constraints, potential attacks on the agent, $Q$-values, and key study terms (in layperson language). Participants were then shown three videos of the RL agent: one where the agent succeeds without an attack, one where the agent fails under attack, and one where the agent succeeds under attack with a safety patch. After each phase, participants answered four questions across two scenarios: one where the agent was under attack without a safety mechanism and one where the agent was patched with a safety mechanism. The first three questions \footnote{The fourth question is discussed in Section \ref{utility-results}} were: (i) Q1. What action will the agent likely take in a state with high risk? (ii) Q2. Why did the agent choose that action? (iii) Q3. Will the agent succeed (reaching the goal safely)? Q1 evaluates participants' basic comprehension of the explanations and the impact of $Q$-values ($Q_{\mathrm{task}}, Q_{\mathrm{risk}}$). Q2 assesses whether participants can identify the primary motivation behind the agent's action at a given state evaluating whether the decision was made to achieve the goal, avoid a risk area, respond to an adversarial attack. This evaluation provides insight into the participant’s comprehension of the agent’s motivations rather than an analysis of the action's overall effectiveness which is measured by Q3. Navigation 2 environment was selected because it provides a sufficient level of complexity that would allow participants to focus on understanding the safety aspects of the agent's behavior rather than being distracted by learning the environment itself. Participants were compensated with a base payment of \$1.5, plus an additional bonus of 10 cents for each correct answer.

\begin{table}[tb!]
\centering
\caption{\textbf{\textbf{H1} Results:} Comparison of the average accuracy of participants' answers across all three safe and three unsafe agents using local, global, xSRL explanation methods.}
\label{explanation baselines}
\begin{tabular}{lcc}
\toprule
\textbf{Explanation Method} & \textbf{Unsafe Agents} & \textbf{Safe Agents} \\
\midrule
Local Explanation       & 48.14\% & 68.88\% \\
Global Explanation      & \textbf{60.37\%} & 73.7\% \\
\textbf{xSRL Explanation}        & \textbf{60.37\%}& \textbf{77.4\%} \\
\bottomrule
\end{tabular}
\end{table}

\begin{table}[tb!]
\centering
\caption{\textbf{H2} Results: Comparison of the average accuracy of the participants' answers for three agents patched with AdvExRL, RRL-MF, and SQRL across all explanation methods.}
\begin{tabular}{lccc}
\toprule
{Explanation Method} & \textbf{AdvExRL} & \textbf{RRL-MF} & \textbf{SQRL} \\ \midrule
{Local Explanation}   & \textbf{74.44\%}      &  68.8\%                & 63.3\%             \\
{Global Explanation}  &    \textbf{80\%}             & 40.2\%                & 73.3\%             \\
\textbf{xSRL Explanation}    &  \textbf{ 81.1\%}              &    78.9\%            &        72.20\%      \\
\bottomrule
\end{tabular}
\label{table:patched}
\end{table}

\vspace{3pt}
\noindent \textbf{Hypotheses.} We will test two hypotheses:

\noindent \textbf{H1:} We hypothesized that using only local or global explanations would be less effective in helping users identify why the agent made certain decisions and whether or not it would succeed under attack, compared to xSRL’s combined approach. In other words, participants presented with local or global explanation methods alone will struggle with Q1-Q3 for both safe and unsafe agents, while those presented with xSRL's explanation will perform better.

\noindent \textbf{H2:} We hypothesized that agents patched with separate safety policies (AdvExRL and RRL-MF) would be more interpretable than agents using joint optimization like SQRL, where the reasoning behind actions under attack is harder to pinpoint.

\noindent \textbf{Results for \textbf{H1}.} Table \ref{explanation baselines} presents the results for \textbf{H1} for both unsafe (under attack agents without safety mechanisms) and safe agents (agents with active safety patches).
Results indicate that safe agents were overall more interpretable across all explanation methods, with xSRL providing the highest average accuracy at 77.4\%. This is likely because participants could observe $Q_{\mathrm{risk}}$ values increasing and then decreasing after actions, which emphasized the agent's safe behavior. In contrast, for unsafe agents, where $Q_{\mathrm{risk}}$ remained high, participants found it more difficult to interpret the agent's reasoning, particularly with local explanations (48.14\%). Notably, xSRL and global explanations achieved similar performance for unsafe agents, both with an average accuracy of 60.37\%. This suggests that the directed graph format used in both methods made it easier for participants to understand the agent’s behavior, especially in high-risk scenarios, by providing a broader view of the agent’s policy. However, xSRL outperformed both global and local explanations for safe agents, confirming that integrating local and global explanations provides a more comprehensive understanding of the agent’s behavior. Overall, xSRL proved to be the \textit{most effective} explanation method, offering more interpretable insights across both safe and unsafe agents to the users, thus confirming H1.

\vspace{3pt}
\noindent \textbf{Results for H2.} The results from Table \ref{table:patched} reveal important insights regarding the interpretability of the different patched agents (AdvExRL, RRL-MF, and SQRL) across various explanation methods (local, global, and xSRL). AdvExRL consistently shows the highest interpretability across all explanation methods, particularly with xSRL (81.1\%). This suggests that its separate safety optimization policy enhances explainability. RRL-MF, while also using a separate safety policy, has slightly lower interpretability (78.9\% with xSRL), likely due to challenges in risk estimation. SQRL, which employs joint optimization of task and safety, demonstrates the lowest interpretability (63.3\% with local explanations), suggesting that the joint approach makes it harder for participants to grasp the agent’s decision-making process, particularly in risk-laden scenarios. This trend reinforces the value of having the safety mechanism separated, as demonstrated by AdvExRL's superior performance across all methods, particularly under attack scenarios, where its safety mechanism is more easily understood by participants. Moreover, global and xSRL explanations overall provided a clearer understanding, specifically for safe agents.


\vspace{3pt}
\noindent \textit{Please refer to Appendix C and D for the detailed  results for Navigation 2 and Maze environments. Appendix E has the details of the studies and their design.}

\begin{figure}
    \centering
    \subfloat[\centering Safety (\%)]{{
        \includegraphics[width=0.44\linewidth]{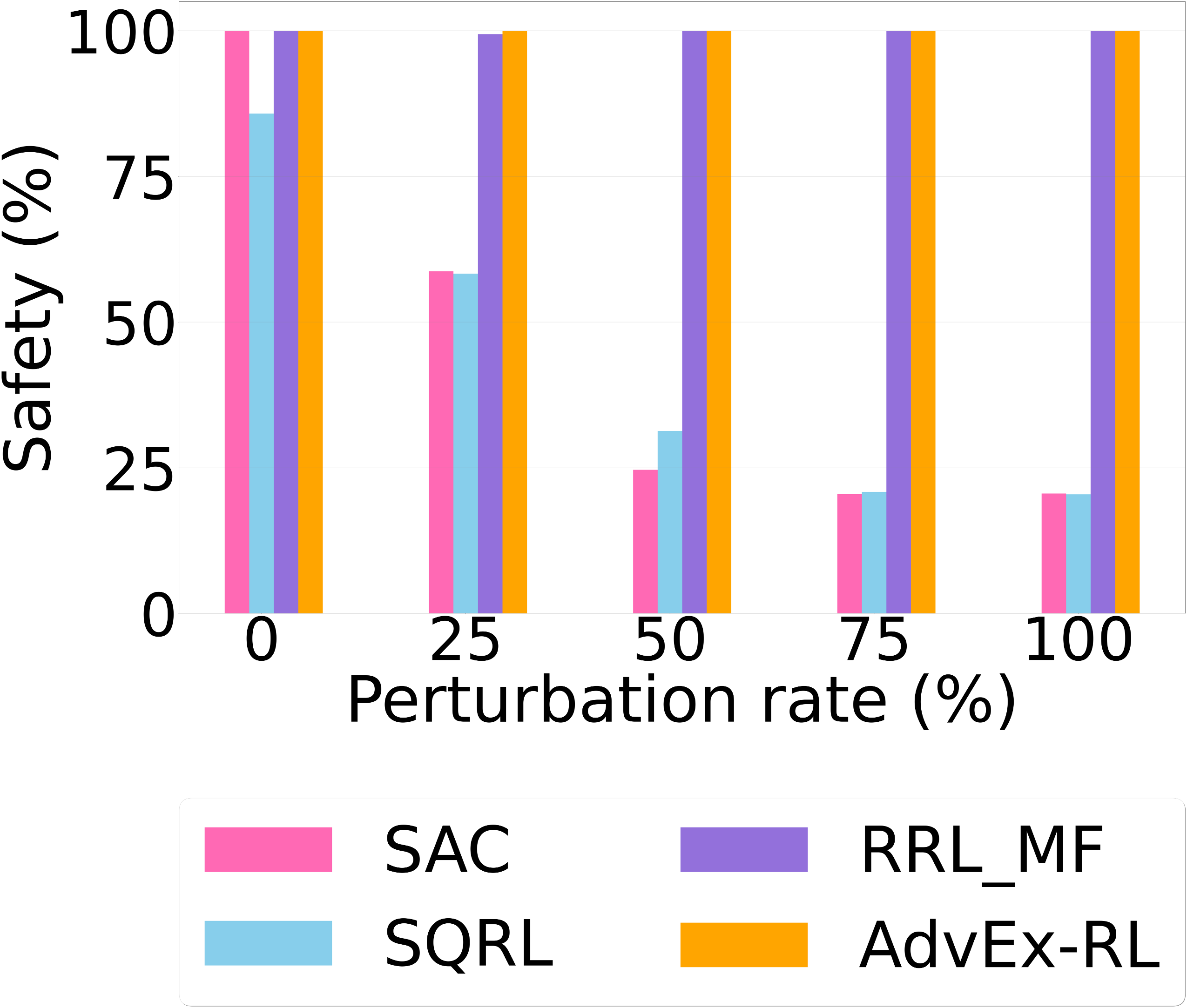}
        }}%
    \qquad
    \subfloat[\centering Success-safety (\%)]{{
        \includegraphics[width=0.44\linewidth]{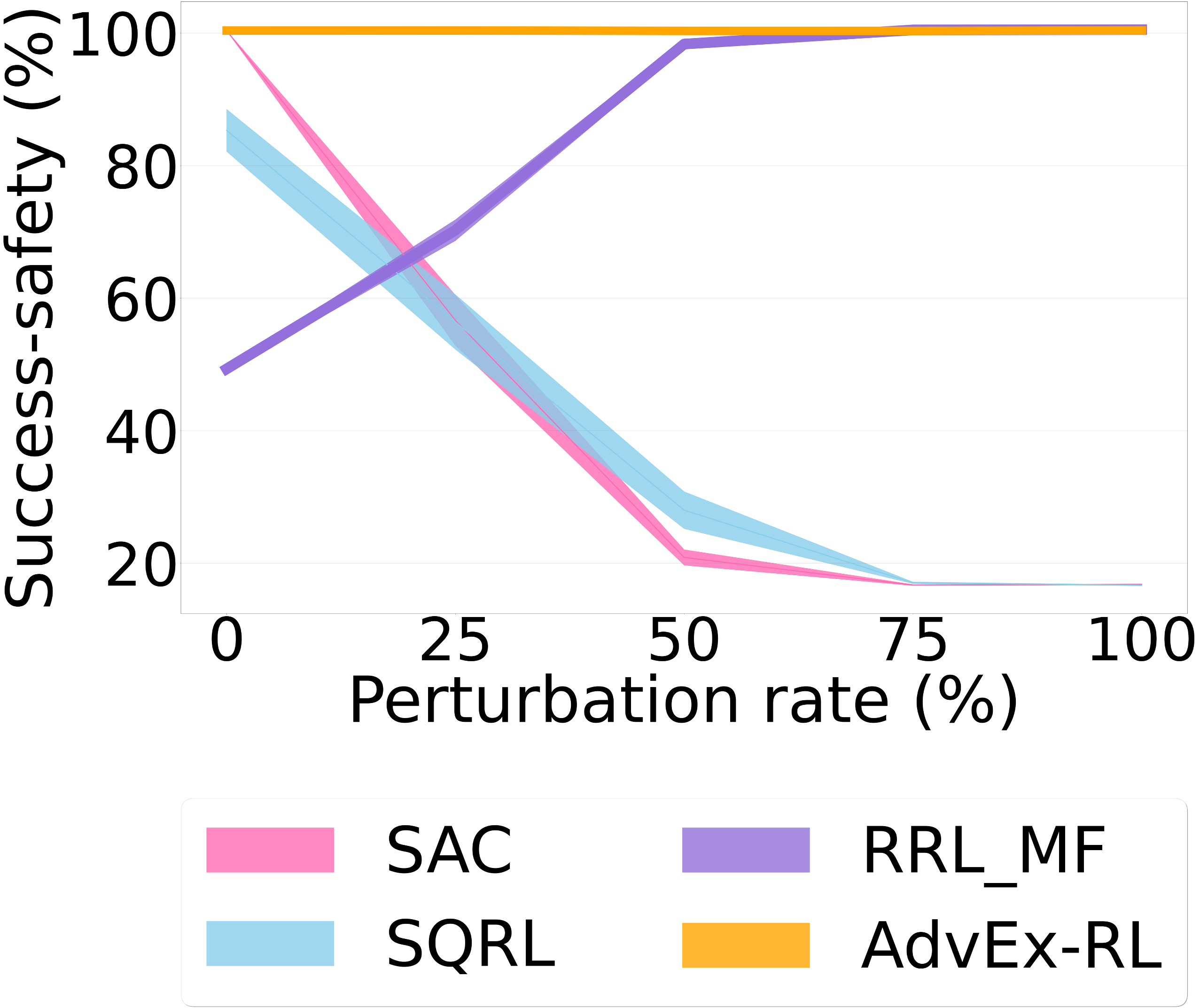} %
        }}%
  \caption{Safety(\%) and success-safety(\%) performance of the patched agents under the influence of various rates of the explanation-guided attack in Navigation 2.}
\label{fig:attack-imapct}
\vspace{-10pt}
\end{figure}

\begin{figure*}[t] 
  \centering
  \begin{subfigure}[b]{0.32\textwidth}
    \centering
    \includegraphics[width=\textwidth]{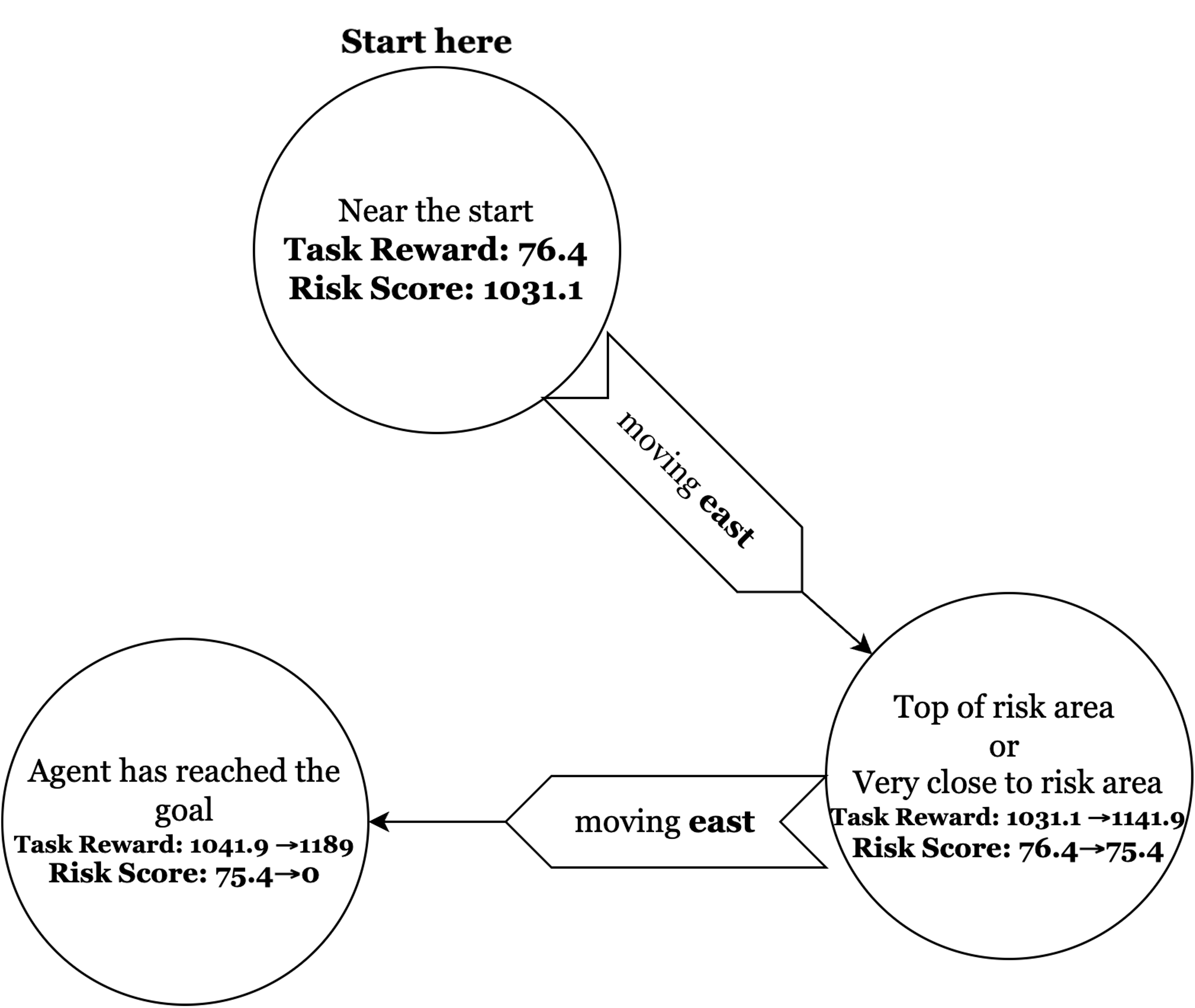} 
    \caption{xSRL explanation of SAC agent \textit{without} attack}
    \label{fig:sac-noattack}
  \end{subfigure}
  \hfill
  \begin{subfigure}[b]{0.32\textwidth}
    \centering
    \includegraphics[width=\textwidth]{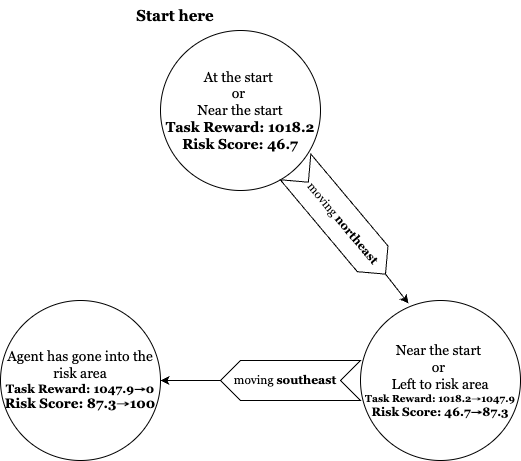} 
    \caption{xSRL explanation of the SAC agent under attack without a safety patching mechanism (therefore, no task or safety decision labels on the edges).}
    \label{fig:sac_attack}
  \end{subfigure}
  \hfill
    \begin{subfigure}[b]{0.32\textwidth}
        \centering
        \includegraphics[width=\textwidth]{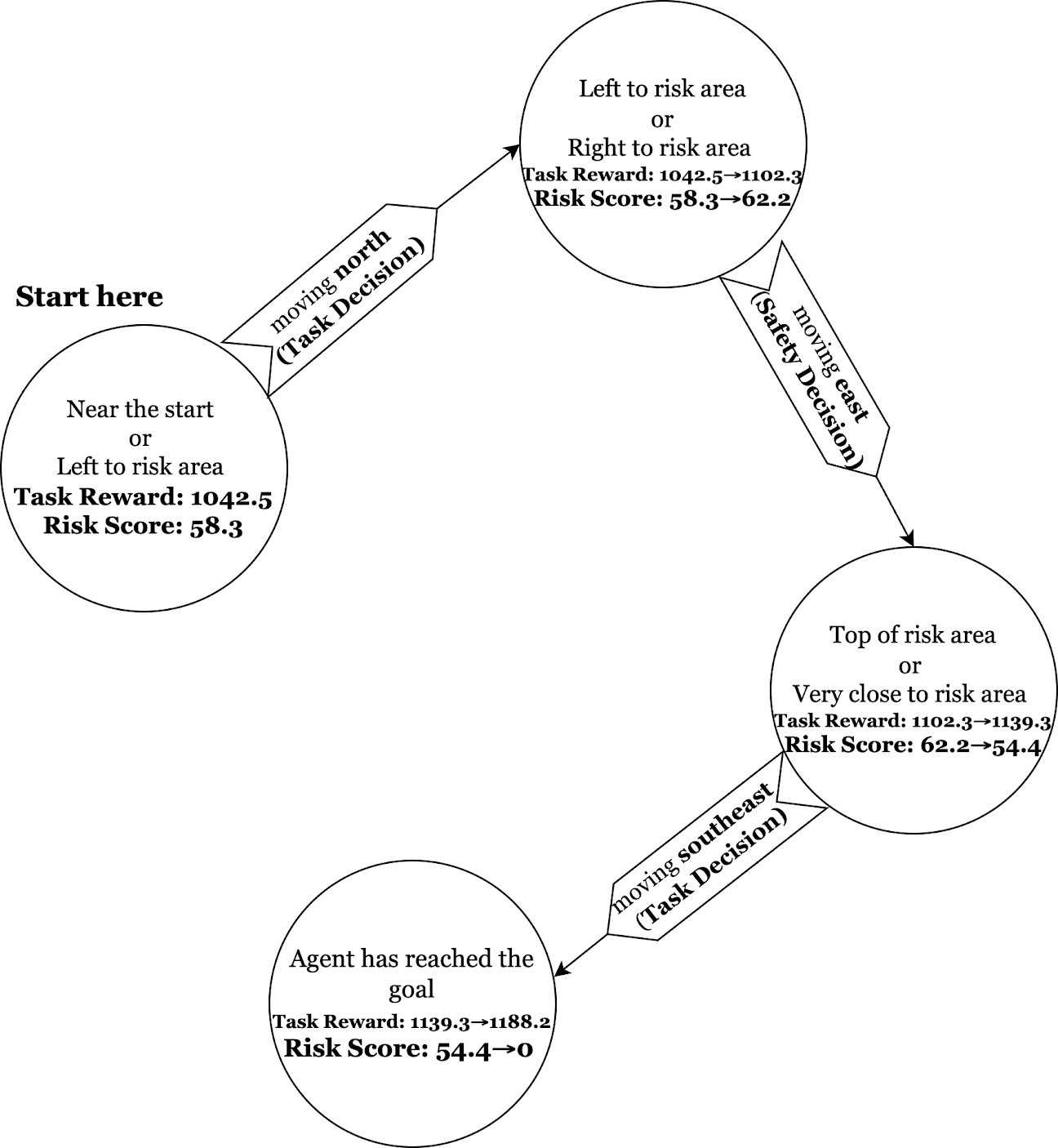}
        \caption{xSRL explanation of SAC \cite{sac} agent under attack with a safety patch}
        \label{fig:sac_advexrl}
    \end{subfigure}
  \caption{An example of using (a) the xSRL explanation graph to launch an attack on the SAC agent at high-risk states, (b) the xSRL explanation of the agent's behavior under attack, and (c) the explanation of the same agent's behavior under the same attack after being patched with the safe policy from AdvExRL \cite{adv}.}
  \label{fig:main_figure}
\end{figure*}

\subsection{Utility of xSRL's Explanations } \label{utility-results}

This section evaluates the utility of xSRL's safety explanations in identifying and resolving vulnerabilities in the agent’s policy using both computational and empirical approach. The computational approach involves (1) assessing the impact of the explanation-guided attack developed in Section \ref{guided-attack}, and (2) measuring the effectiveness of patching techniques in improving the agent's safety under the same attack. The empirical approach involves analyzing the final question in our user studies to determine whether participants can distinguish between safe and unsafe agents.

\subsubsection{Impact of Explanation-guided Attack and Patching Techniques.} The explanation-guided attack (AAA) is implemented by targeting abstract states that exhibit higher $Q_{\mathrm{risk}}$ values in the xSRL graph of the SAC agent (Figure \ref{fig:sac-noattack}). We attacked 0, 25, 75 and 100\% of the concrete states clustered within these high-risk abstract states. Figure \ref{fig:sac_attack} shows the behavior of the SAC agent under attack, without any safety mechanism. To measure the attack’s impact, we used two metrics from \cite{adv}, including \textit{Safety(\%)}, which quantifies the proportion of time the agent acts safely over its maximum episode length, and \textit{Success-Safety(\%)}, which indicates how close the agent is to reaching its goal. An agent is deemed successful if it finishes the episode within a predefined minimum distance from the goal. For the distance threshold, we used the same value specified in \cite{adv} for Navigation2 environment. Figure \ref{fig:attack-imapct} presents the impact of the attack on SAC without a safety mechanism, as well as patched agents using AdvExRL, RRL-MF, and SQRL, all under 0-100\% attack rates. The results clearly show that SAC suffers the most, as its vulnerability is evident in the xSRL graph for its behavior (Figure \ref{fig:sac_attack}). In contrast, AdvExRL is the safest, as indicated by the minimal impact of the attack on its safety and success, which is also visually reinforced by its xSRL explanation graph (Figure \ref{fig:sac_advexrl}). Please refer to Appendix C for all remaining results for the agents under attack.



\subsubsection{User Studies.} In our previous user studies (Section \ref{sec:trust}), we included a subjective question in which participants were asked to identify which agent is safer among two agents, vulnerable and patched. Participants were also asked to rate their confidence in their choice on a Likert scale from 1 (``not confident at all'') to 5 (``very confident''), and we reported the responses with confidence levels of $\geq 4$. Both agents were subjected to identical attack scenarios, but \textit{only one} of which was patched with a safety mechanism. To minimize learning effects, participants were allowed to switch between the scenarios and adjust their answers at any point during the study. We hypothesize that if participants had a correct mental model of the agents’ strategies, they would be able to identify the safer agent.

Our results show that both xSRL and global explanations achieved comparable results across all patching methods (Table \ref{safer-agent}). This is likely due to these methods' ability to provide a comprehensive view of the agent's behavior from the start state to the goal state. This allowed participants to observe the final outcomes (success or failure), which directly influenced their choice of the safer agent. Interestingly, despite the results of the objective questions in Table \ref{table:patched} showing that AdvExRL agents were more interpretable, participants expressed higher confidence when evaluating RRL-MF agents in the subjective questions. Overall, participants were able to recognize the safer agents with high confidence across all explanation methods, with confidence levels exceeding 50\% in most cases (except for the local explanation of AdvExRL, which was lower). These results reinforce the importance of providing clear and comprehensive safety explanations to aid in understanding behavior of RL agents, particularly in safety-critical environments in which our approach, xSRL, proved to be effective.

\begin{table}[tb!]
\centering
\caption{\% of subjects selecting the safer agents across the nine conditions, with confidence ratings (${\geq}4$) reported in parentheses.}\label{safer-agent}
\begin{tabular}{@{}l*{3}{>{\centering\arraybackslash}p{6em}}@{}}
\toprule
& \textbf{Local} & \textbf{Global} & \textbf{xSRL} \\
\midrule
\textbf{AdvExRL} & 40\%(52.6\%) & 83.33\%(62.9\%) & 70\% (54.17\%) \\
\textbf{RRL-MF} & 93.33\%(72.4\%) & 90\% (44.83\%)& 90\%(62\%) \\
\textbf{SQRL} & 76.67\%(59.2\%) & 80\%(42.86\%) & 83.33\%(44.44\%) \\
\bottomrule
\end{tabular}
\vspace{-10pt}
\end{table}

\section{Conclusion}
In this paper, we introduced xSRL, a framework that combines local and global explanations to address safety constraints in reinforcement learning (RL) agents. Our approach aims to improve the interpretability of RL policies, with a focus on safety-related behavior, offering developers insights to identify and address vulnerabilities in agent policies. By utilizing policy abstraction and safety-specific visualizations, xSRL provides a clearer understanding of agent decisions and helps users refine policies based on safety concerns. Through computational and empirical evaluations, we showed that xSRL enhances both trust and utility, offering a useful tool for RL policy testing and refinement in safety-critical environments. These results highlight the value of safety-aware explanations in supporting more effective RL system development and user interaction.

\bibliographystyle{ACM-Reference-Format}
\bibliography{main}


\end{document}